\documentclass[11pt]{article}

\usepackage[final]{acl}

\usepackage{times}
\usepackage{latexsym}

\usepackage[T1]{fontenc}
\usepackage[utf8]{inputenc}

\usepackage{microtype}

\usepackage{inconsolata}

\usepackage{graphicx}

\usepackage{booktabs}
\usepackage{multirow}
\usepackage{float}
\usepackage{enumitem}
\usepackage{amsmath}
\usepackage{url}
\usepackage{tipa}

\title{One Form to Transfer Them All: Pretraining Multilingual Language Models Beyond Native Orthography}

\author{
Muge Zhang$^{1}$ \quad Aaron Jencks$^{1}$ \quad Krishna Badikela$^{1}$ \\
\textbf{Yulia Tsvetkov}$^{2}$ \quad \textbf{Sachin Kumar}$^{1}$ \\
$^{1}$Department of Computer Science and Engineering, Ohio State University \\
$^{2}$Paul G. Allen School of Computer Science \& Engineering, University of Washington \\
\texttt{contact: \{zhang.16414,kumar.1145\}@osu.edu}
}

\begin{document}
\maketitle

\begin{abstract}
Multilingual language models transfer knowledge across languages through shared subword vocabulary, a mechanism that breaks down when related languages use different writing systems. Prior work addresses this via script equalization (romanization or IPA transcription), but direct comparisons are rare; the focus has been on encoder-only models, with most work adapting existing pretrained models. 
% finetuning-only, or non-controlled setups. We treat the input representation as a first-class variable in 
We systematically compare different input representations in autoregressive multilingual pretraining, comparing orthographic text, IPA, and romanization in a controlled setup across three scales (467M, 709M, and 1.03B) on eight languages in four typologically motivated pairs. %We evaluate under prompting and finetuning on classification and generative tasks, including seen and unseen languages. 
Across a wide range of downstream tasks on seen and unseen languages, romanized pretraining yields the strongest cross-lingual transfer, and the advantage over text widens with scale. IPA improves over text in most settings but trails romanization. Surprisingly, finetuning a text-pretrained model on romanized data hurts performance on languages already covered by the base model, only marginally helping when the model lacks script coverage. %; controlled experiments show it only helps when the base model lacks script coverage. 
Our results indicate that for multilingual models spanning typologically diverse scripts, to obtain maximum benefits, romanization should be treated as a core design choice applied at pretraining rather than a post hoc fix.
% 's benefits are thus best captured at pretraining time rather than as a downstream conversion.
Our code is available at \url{https://github.com/skai-research/one-form-transfer}.
\end{abstract}

\section{Introduction}
In multilingual language models (LMs), vocabulary overlap is the primary source of cross-lingual transfer among languages
% Multilingual language models (MLMs) transfer knowledge across different languages primarily through shared subword tokens 
\citep{conneau2020, bigscience2022bloom, ustun2024aya}. 
When two languages do not share writing systems, this overlap is largely impossible, even for linguistically close languages as the surface form of the text obscures the similarities \citep{wu2020languages,muller2021unseen,stap-etal-2023-viewing}. %as those languages will not share any string overlap
% When two languages are written in the same script, linguistic transfer happens naturally through cognates and borrowed words mapping to overlapping vocabularies entries and sharing embeddings. 
% When they do not share script, however, the same mechanism breaks down, even for languages that are linguistically very close. 
% In these cases, the surface form of the text obscures linguistic similarities that the model can otherwise exploit, and prior works have identified script mismatch to the dominant failure mode of linguistic transfer \citep{muller2021unseen, wu2020languages} and, more broadly, cross-linguistic performance gaps largely reflect representation and design choices rather than intrinsic linguistic difficulty\citep{shani2026roots}.

\begin{figure*}[t]
\centering
\includegraphics{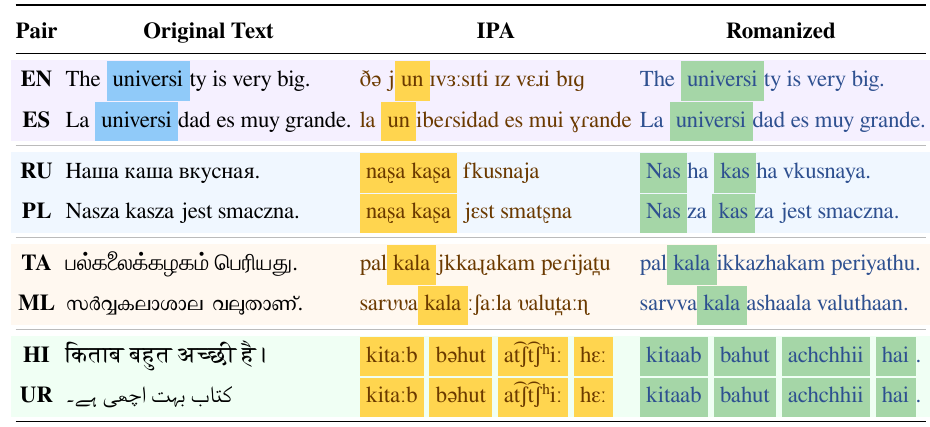}
\caption{Example sentences across four language pairs in original script, IPA, and romanized form. Shared subword tokens are highlighted: {blue} for original text, {gold} for IPA, and {green} for romanized. %\mz{not sure if its ok to make the figure this small. if not, maybe we could show some in text examples rather than a figure to illustrate this?}\sk{not against it but the font is too small now though, anyway to make it larger?} \mz{now font is around 50\% larger}\sk{increase legend too and we should be good, also can you center the legend or perhaps remove the length and mention the color in caption}\mz{removed legends. }
}
\label{fig:alignment-example}
\end{figure*}

% One response to this problem is to convert text in script neutral format. 
% \sk{figure 1 could be cited somewhere in this paragraph.} \mz{added at the beginning}
To increase cross-lingual overlap, prior work has explored different methods of \emph{script equalization}---converting all languages to a shared form (see \autoref{fig:alignment-example}). This is done primarily through two mechanisms. First is romanization \citep{hermjakob-etal-2018-box}, where non-Latin scripts are mapped to Latin characters using transliteration tools \citep{purkayastha2023romanization, husain2024romansetu, ebing2026onescript}. %\sk{if needed for space, could keep only a couple of citations here and discuss rest in related work.}. \mz{modified}
%\sk{are there another romanization tools we can cite? to keep the intro more general.}. %such as Uroman\citep{hermjakob-etal-2018-box}, exposing the similarities between languages that their original orthographies hide. 
Second is phonemization, where text is mapped to phonemic representations such as IPA \citep{international1999handbook}, where words that sound alike across different languages would share forms \citep{nguyen2023phonexl,goriely2024babble, jung2024mitigating, goriely2025ipa}. %\sk{more IPA-LM citations? i remember there's atleast a couple more -- David mortensen has one.}. \mz{added}
% A growing body of work has used these representations to improve cross-lingual transfer 
However, these works have largely focused on improving transfer during \textit{finetuning} with off-the-shelf encoder-only models pretrained with orthographic text, almost exclusively on English. Furthermore, both mechanisms have been explored separately but not compared in modern language models, which are dominantly autoregressive.
% these works mostly adapt English or near-English pretrained models, romanization is applied only at fine-tuning, and no other representation is compared under a controlled setup. 
% They leave open two questions: whether the benefit of these representations is best captured at pretraining or finetuning, and how do the two mechanisms compare.

In this work, we focus on \textit{pretraining autoregressive} multilingual LMs, 
% we treat the input representation as a first-class variable in multilingual pretraining and
comparing both script equalization mechanisms in a controlled setting with orthographic text. %  three forms of the same text: native orthographic text, Uroman romanization \citep{hermjakob-etal-2018-box}, and phonemic transcription in the International Phonetic Alphabet (IPA). 
We pretrain models from scratch on an eight-language corpus organized into four typologically motivated pairs---English--Spanish, Russian--Polish, Hindi--Urdu, and Tamil--Malayalam---chosen to vary the relationship between orthography and phonology. We train at three scales (467M, 709M, and 1.03B parameters) %\sk{maybe just add the medium scale since there's only 3} \mz{done} 
with each representation, holding architecture, data, vocabulary size, and training procedure constant.
We evaluate the trained models under prompting on pretraining-seen languages and under fine-tuning on both seen and unseen languages, covering classification and generative tasks.

We find that romanized pretraining is the strongest configuration in every evaluation regime and at every model size, with the gap from text widening as parameters scale. IPA improves over orthographic text in most settings and matches romanization in the narrow case of Hindi and Urdu, the pair in our corpus that is most phonologically aligned and most orthographically disjoint, but lags romanization on transfer to unseen languages. %\sk{could add a sentence here or maybe in the results later citing other IPA works who do not show improvements}.\mz{added in the results section. do we still want to mention the same thing in the intro?} 
Surprisingly, the recipe of romanized finetuning applied to a text-pretrained model, reported as beneficial in prior work, instead regresses performance on the languages a multilingual model already covers; we reproduce the prior gains on a controlled English-only setup and show that the recipe helps only when the base model lacks script coverage in the first place. Together, these results indicate that romanization helps by aligning input with the lexical structure the model has internalized during pretraining; the gains it delivers are therefore most effectively captured by adapting romanization at pretraining rather than downstream. % a downstream conversion step. \textcolor{red}{
 Our findings suggest that future work on multilingual modeling should treat the input representation as a deliberate design choice, with romanization as a strong default when transfer to unseen scripts is a priority.%} %\sk{will come back later but should end with some sort of recommendations or something. -- like our work provides guidelines for future work in this area. %} \mz{added}

\section{Related Work}

\paragraph{Script barrier in multilingual LMs.}

Multilingual pretrained models \citep{devlin2019bert,conneau2020,bigscience2022bloom,aryabumi2024aya23, xue-etal-2021-mt5, lin-etal-2022-shot, grattafiori2024llama3, qwen2024qwen25, yang2025qwen3} %\sk{cite aya,llama3.2,qwen2.5/3,a couple more} \mz{added. also added XGLM and mt5}
transfer well across languages that share scripts but show little improvement or even degrade if not \citep{wu2020languages,lauscher2020zero}. Prior studies identify script mismatch, and hence lack of vocabulary overlap, as the dominant failure mode of transfer to unseen languages \citep{pires-etal-2019-multilingual,muller2021unseen,stap-etal-2023-viewing}. Our work targets this barrier by changing the input representations. % rather than the model. %\textcolor{red}{Multilingual pretrained models such as mBERT \citep{devlin2019bert}, XLM-R \citep{conneau2020}, BLOOM \citep{bigscience2022bloom}, and Aya \citep{ustun2024aya, aryabumi2024aya23} transfer well across languages that share scripts but degrade when source and target languages use different scripts \citep{wu2020languages, lauscher2020zero}.}\sk{these papers are somewhat old and are encoder only models. I would atleast cite BLOOM, and AYA, there are many more models that discuss scripted related issues.}\yt{+1}\mz{added}
Prior work has also shown that amount of vocabulary overlap is
not correlated with better performance \citep{meyer2024systematic, limisiewicz2023tokenization, kk2020crosslingual}. Our results echo these findings: improvements do not track overlap, and, as an example, English–Spanish transfer holds up under IPA even as their shared vocabulary becomes smaller. %[something about our improvements are not correlated with overlap amount, and also, despite drop in english/spanish overlap our performance doesn't drop.]\sk{todo}
%\aj{do we need these last two sentences at all? I think we could just drop them.}\sk{we are claiming that overlap is necessary so presenting a counterargument that it might not be the only factor at play is part of related work.}% \mz{}

\paragraph{Romanization for cross-lingual transfer.}

A growing body of work has used romanization to expose lexical overlap that the original orthography hides. Prior work on encoder-only models has shown that transliteration, or romanization, can substantially benefit multilingual learning, particularly for low-resource languages, both in Indic settings and across broader multilingual corpora \citep{purkayastha2023romanization, moosa2023does, ebing2026onescript, jung2026happiness}. More recent work has extended these ideas to decoder-only LMs through continued pretraining and instruction tuning on romanized text \citep{husain2024romansetu}, contrastive alignment \citep{liu2024translico}, and tokenizer adaptation for transliterated input \citep{liu2025transmi}. We instead pretrain decoder-only autoregressive LMs from scratch on romanized text, performing controlled comparisons with orthographic text to isolate when romanization helps and when it hurts across both finetuning and prompting regimes. % \aj{How's this?}

\paragraph{Phonemic representations.}

A parallel line of work has explored phonemic input as another way to bridge scripts. \citet{nguyen2023xphonebert} introduced XPhoneBERT, a multilingual phoneme-level encoder trained on phoneme sequences from nearly 100 languages for speech-related applications; \citet{jung2024mitigating} showed that phoneme-based models reduce the cross-lingual performance gap; \citet{goriely2024babble} pretrained a GPT-2-scale model on phonemized English and found minimal cost on meaning-level tasks. Concurrent work \citep{anonymous2026phonemes} compares Text vs. IPA subword tokenizers across 24 languages and pretrains a 240M model, finding IPA improves text compression without changing task performance. Our work extends this study to include romanization, echoing their findings on compression but also find improved downstream performance with both IPA and romanization over text, contrary to their findings.

% \section{Methodology}
% We study whether phonemized text representation would improve cross-lingual transfer in language models compared to standard orthographic text representation. We first pretrain and then finetune language models under three text representation setups: orthographic text, IPA transcription, and romanization. We keep all other variables including architecture, data, training procedures constant to maintain a controlled comparison so the effect of input representation on multilingual generalization is isolated. 

% The IPA representation maps all languages into one shared phonimized symbol space so words that sound similar across different languages have similar transcriptions regardless of which script they were originally written in. For example, loan words that are spelled differently in Cyrillic and latin will converge to IPA sequences that are almost the same. Romanization, by contrast, maps text into Latin characters while maintaining the approximate orthographic pattern rather than the phonemic content, providing a third point of comparison. 

\section{Experimental Setup}
\label{sec:experiment}

\subsection{Language Pairs}
We study eight languages organized into four typologically motivated pairs that span a range of orthographic and phonological relationships (see examples in \autoref{fig:alignment-example}).%\mz{Abraham mentioned that there's a lot more that can be said about these languages that can give more information to the readers. But im a bit undecided about adding all these details given we do not see a strong corelation between language pair overlap and performance.}
% \sk{like what?}\sk{you already have that explanation in the intro, could add it here.}\mz{e.g. giving a more detailed description of script closeness, reasons for choosing them, etc}\sk{there's no space for that, we could do that in the appendix and point to the figure here.}\mz{added figure reference. maybe adding an appendix later for detailed rationales, etc}:
\begin{enumerate}
    \item \textbf{English--Spanish}: Both are Latin-scripted with shared cognates and borrowed words.
    \item \textbf{Russian--Polish}: Different scripts (Cyrillic vs.\ Latin), but related Slavic languages with higher phonological overlap.
    \item \textbf{Hindi--Urdu}: Different scripts (Devanagari vs.\ Arabic), but are mutually intelligible in their spoken form (often considered variants of a single language).
    \item \textbf{Tamil--Malayalam}: Different scripts but from the same language family (Dravidian), with high phonological similarity. 
\end{enumerate}
We choose the last three pairs because they provide varying degrees of phonological and phonetic overlap, and stand to benefit via cross-lingual transfer from shared representations. Pairs such as Hindi--Urdu have minimal orthographic overlap but extensive phonemic overlap, so any benefit from a phonemic representation should be especially visible in such cases.
%\sk{add a line about english/spanish being the control group}.\mz{added}
English--Spanish serves as a control where text representation already provides shared subword tokens, where phonemic representations might reduce overlap. We train a single multilingual model jointly on all eight languages; the four pairs serve to organize our analysis and to anchor finer-grained comparisons in \autoref{app:bilingual-downstream}.% \sk{cite appendix}\aj{cited}

\subsection{Pretraining Dataset Construction}
\label{sec:dataset_construction}
% We construct the pretraining corpus in two stages. 
For each language pair, we subsample monolingual documents from FineWeb-2 \citep{penedo2025fineweb2}, maintaining the word count ratio.\footnote{We use white-space separated tokens as a proxy for words. All of our pretraining languages use white space as delimiters.} Each pair-level corpus is sized to match OpenWebText \citep{gokaslan2019openwebtext}. This procedure preserves natural within-pair resource imbalance while equalizing budget across pairs. %, ensuring that lower-resource pairs (e.g., Tamil--Malayalam) are not overwhelmed by higher-resource ones (e.g., English--Spanish). %\sk{words are not actually computed right? it is just space separated? might be worth adding that here or in a footnote.} for each pair to maintain their relative resourcedness.\mz{footnode added.}  
Within each pair, we randomly sample a subset for validation ($\sim$3.2M words per pair with the same ratio).  The four bilingual corpora together form our 8-language corpus with $\sim$21.7B words in total. After tokenization, the word-matched corpus yields 50B Text tokens versus 33B IPA/Romanized tokens, putting IPA/Romanized models at a compute disadvantage and making their gains potentially conservative under FLOP-matched training.%}\sk{mention total size in words or in tokens after tokenization?}.\mz{added total word count}  %, the open reconstruction of GPT-2's pretraining corpus ($\sim$8B tokens of English), so every pair receives roughly GPT-2-scale pretraining data. The full eight-language corpus is the concatenation of four such pair-level corpora, resulting in roughly four times the OpenWebText size in total.% All datasets will be released on HuggingFace \citep{XXX}.

\subsection{Text Representations}
We compare the following setups. In the first three, we pretrain/finetune with the same representation.
\paragraph{Orthographic text.} The original text in each language's native script. For language pairs that share a script (English--Spanish, both Latin), there will be natural token overlap. On the other hand, for language pairs with different scripts (e.g., Russian–Polish, Cyrillic vs. Latin), we expect little overlap, limiting transfer. % through shared subwords.

\label{sec:g2p}
\paragraph{IPA transcription.} We convert the corpus into IPA using the Phonemizer library \citep{Bernard2021}.\footnote{We use Phonemizer for its speed and broad language coverage. Alternative phonemizers such as Epitran \citep{mortensen2018epitran} and neural grapheme-to-phoneme models \citep{peters2017massively} offer higher per-language accuracy but are much slower and become a bottleneck at our corpus scales. Future work may revisit this quality-throughput tradeoff as faster tools become available.} To further improve cross-lingual overlap, we remove certain diacritics from the transcriptions, including stress marks (\textipa{\textprimstress}, \textipa{\textsecstress}) and length marks (\textipa{\textlengthmark}), tone accents, and nasalization such that only phonemic segments remain. This logic makes the representation more coarse-grained, increasing overlap.\footnote{Retaining these diacritics in early experiments yielded roughly 50\% the cross-lingual overlap (weighted Jaccard) of the stripped version, motivating this choice.} %, which can be seen in \autoref{fig:tokenizer-overlap} with additional details in \autoref{app:overlap-details}.%\sk{is there old numbers on overlap with the stress markers? or we could say in our early explorations, IPA was terrible or something in the footnote.} \mz{added footnote.}\aj{added reference to the figure with overlap analysis, it will need to be expanded to include ipa full and ipa stripped data from the spreadsheets} %further collapsing language specific structures that will hide underlying phonemic structure. 
We additionally modify the Phonemizer library to preserve words containing digits or characters outside the source language's script, which the default pipeline transcribes ambiguously.

\paragraph{Romanization.} We romanize all texts using the Uroman library \citep{hermjakob-etal-2018-box}.\footnote{We use Uroman because it covers almost all scripts with a rule-based system; \citet{purkayastha2023romanization} find it performs comparably to or better than language-specific transliterators.} For languages already in Latin (e.g., English, Spanish, Polish), the romanized form is close to the original but not always identical.\footnote{English text passes through unchanged (e.g., \emph{The quick brown fox} $\rightarrow$ \emph{The quick brown fox}), whereas Spanish and Polish lose diacritics and special characters (e.g., Spanish \emph{el niño está} $\rightarrow$ \emph{el nino esta}; Polish \emph{Łódź} $\rightarrow$ \emph{Lodz}).} %\mz{latin-script romanization examples added in footnote. }

\paragraph{Text-to-romanized (Text$\rightarrow$Rom).} In this setup, we finetune our \textit{text}-pretrained checkpoints on romanized downstream task data, mirroring the standard recipe in prior work \citep{husain2024romansetu, purkayastha2023romanization}. This tests whether romanization's benefits can be recovered downstream, without the cost of romanized pretraining.\footnote{We omit Text$\rightarrow$IPA because IPA has essentially no vocabulary overlap with the orthographic pretraining data; therefore, the pretrained token embeddings provide little useful initialization for IPA tokens.}%\sk{should we say something in the footnote about why not text->IPA? because of no expectation of overlap?}

\subsection{Tokenization}

\label{sec:tokenization}

For each representation, we train a single Byte-Level BPE tokenizer \citep{sennrich2016bpe,radford2019gpt2} jointly on the eight-language corpus using the HuggingFace tokenizers library \citep{huggingface_tokenizers}.\footnote{After the paper submission, we discovered an integer-overflow bug in the BPE training library. Our analysis in \autoref{app:i64} shows that its effect on our results are negligible and does not alter any conclusions.} We use a vocabulary size of 100K to accommodate the combined character inventory across the eight languages. Keeping vocabulary size constant across representations ensures a controlled comparison in which the only variable is the surface form of the text.

\paragraph{Tokenizer overlap analysis.}

Before pretraining, we quantify the potential for cross-lingual transfer at the lexical level by measuring how much of each tokenizer's vocabulary is shared between languages. Two languages tokenized into largely disjoint token sets cannot easily share embedding-level representations with limited transfer. Therefore, 
we expect orthographic tokenizers to show high overlap only for script-sharing pairs (English--Spanish), while IPA and romanization are expected to produce higher overlap across similar languages by collapsing scripts into a common symbol set. 
We report a corpus-size-normalized, frequency-weighted Jaccard overlap between the unique tokens observed in each language's monolingual corpus; full details, including the corrections we apply for rare tokens, punctuation artifacts, code-switching, and corpus-size imbalance, are in \autoref{app:overlap-details}.

\paragraph{Sequence length analysis.}Another consequence of representation choice is how compactly each language's text is encoded. Sequence-length differences across languages are a source of unfairness in multilingual models. Languages that tokenize into more tokens per document require higher training compute, inference latency, per-token API cost, and consume more of the model's effective context window \citep{petrov2023language, ahia2023languages}. We therefore measure token sequence length for the three tokenizers as the average number of tokenizer output tokens per document on the FLORES parallel corpus \citep{nllb2022nollanguageleftbehind}, which expresses the same content across all eight languages and therefore makes per-document token counts directly comparable across languages.

\subsection{Pre-training}
For each representation, we pretrain a causal language model from scratch. We use a modified version of NanoGPT, \texttt{modded-nanogpt} \citep{modded_nanogpt_2024}, which incorporates a number of training efficiency improvements over the original \citep{Karpathy2022}, obtaining better performance. % and reaches lower FineWeb validation loss than NanoGPT at the same compute.
\paragraph{Architecture.}
We train models at three scales to study the interaction between model size and representation, summarized in \autoref{tab:model-sizes}. Following \texttt{modded-nanogpt}, each model uses value embeddings, which are three additional vocabulary-sized tables whose outputs are mixed into the value projection of the first and last three attention blocks via learned scalars, inspired by the value-residual connections of \citet{zhou-etal-2025-value}. These contribute negligible FLOPs but a large share of parameters; % (49\%, 43\%, 37\% of totals at Small, Medium, Large); 
we treat non-embedding parameters as the basis for cross-scale comparison.

\begin{table}[h]
\centering
\small
\setlength{\tabcolsep}{4pt}
\begin{tabular}{lccc}
\toprule
\textbf{Size} & \textbf{L / H / $d_{\mathrm{model}}$} & \textbf{Params} & \textbf{Non-emb.} \\
\midrule
Small  & 12 / 6 / 768   & 467M  & 83M  \\
Medium & 16 / 8 / 1024  & 709M  & 197M \\
Large  & 20 / 10 / 1280 & 1.03B & 387M \\
\bottomrule
\end{tabular}
\caption{Model configurations. Each is trained for all three representations.}
\label{tab:model-sizes}
\end{table}

\paragraph{Training procedure.}

All models use a shared optimization setup and checkpoint selection procedure to ensure controlled comparisons across representations. %\aj{I've copied the paragraph to the appendix and had ChatGPT extract the most important sentence.}            
We use a dual-optimizer setup: AdamW ($\beta_1{=}0.8$, $\beta_2{=}0.95$) for embedding, head, and scalar parameters, and Muon (momentum${=}0.95$) for hidden-layer matrix parameters, with a linear cooldown learning-rate schedule (peak Muon learning rate as $0.025$). Each training step processes 524{,}288 tokens across 8 NVIDIA H100 GPUs. Full training details are provided in \autoref{app:pretraining}. %\sk{how long do you train for? we could also include training/validation logs in the appendix.} \mz{added training time}
% We train each model for 3 epochs over the training split of the pretraining corpus, and select the checkpoint with the lowest validation loss on the held-out split for downstream evaluation.\footnote{All models in this study are trained for a fixed token budget that is below the compute-optimal point for their parameter count \citep{hoffmann2022training}; but the controlled comparison across representations is preserved since the budget is held constant.}
%\sk{we can move this paragraph to appendix and just say one line here referring to it if needed for space.}

\paragraph{Bilingual reference models.}
As controlled references that isolate representation effects on individual language pairs, we additionally train bilingual models on each of the four pairs. These models and their analyses are not part of the main results; we describe them and report their behavior in \autoref{app:bilingual-downstream}, where they primarily serve as motivation for the multilingual setting that is our main focus.%\sk{marked as potential for appendix?} \aj{I can't really get this any shorter, unless you mean to completely drop this paragraph}

\subsection{Evaluation}
We evaluate the models under two regimes: directly prompting the pretrained checkpoints, and supervised fine-tuning on classification and generation tasks. In both settings, we evaluate on languages that were seen during pretraining. To test whether representation effects generalize beyond the training distributions, we finetune on select unseen languages%\sk{i think should mention names of unseen languages here, saying that they are covered by different evaluations, not by each} \mz{added. realizing we dont have unseen ICL. i'll try to get the results for unseen ICL before submission. should not take too long.} \mz{jobs runnnig now}
, which include Arabic and French (which share scripts with the pretraining languages) as well as Bengali and Greek (which do not). %\mz{modified description about seen/unseen}% which are covered by different evaluations, not by each.

\paragraph{Prompting.}
We score each pretrained model under zero-shot and few-shot setups on two benchmarks: XStoryCloze \citep{lin-etal-2022-shot} and XCOPA \citep{ponti-etal-2020-xcopa}. %\sk{this detail about appendix could be mentioned in the results section when you discuss implications.} \mz{moved to 4.2}. 
Each benchmark covers a subset of our pretraining languages along with a set of unseen languages, which lets us probe for transfer. Within each \textit{$\langle$task, language, representation$\rangle$} cell, we score every candidate label by likelihood and predict by argmax: per-token-average log-likelihood for label-word tasks (normalized by the number of candidate tokens) \citep{zhao2021calibrate} and sum log-likelihood for completion-style multiple choice \citep{brown2020gpt3, lin-etal-2022-shot}. Label words and structural scaffolding are localized to each cell so that all three representations evaluate identical underlying inputs, and the model never sees out-of-representation tokens mid-prompt.

\paragraph{Fine-tuning.}
We finetune the pretrained checkpoints on three downstream tasks: natural language inference \citep[XNLI; IndicNLI,][]{conneau-etal-2018-xnli,aggarwal-etal-2022-indicxnli}, multilingual intent classification \citep[MASSIVE,][]{fitzgerald-etal-2023-massive}, and multilingual abstractive summarization \citep[XL-Sum,][]{hasan-etal-2021-xl}. The classification tasks jointly cover all eight pretraining and the unseen languages; For XL-Sum, we finetune on English, Spanish, Russian, Hindi, Urdu, and Tamil (other languages were not available). For each \textit{$\langle$task, representation, scale$\rangle$} cell, both inputs and references (including target summaries) are converted into the corresponding representation using the same Phonemizer and Uroman pipelines as in pretraining. We report macro-F1 for classification tasks and ROUGE-L on XL-Sum. Hyperparameter search and other training details are in \autoref{app:finetuning-details}.%\sk{include in appendix that seen languages are jointly trained and unseen individually?}\mz{added}%\sk{could move some grid search etc details to appendix if needed for space.} \mz{moved to appendix}%\sk{add a line about the references also being transcribed}.\mz{added}

\section{Results and Discussion}
\label{sec:results}

We organize our results into two parts. We first report cross-lingual subword overlap and sequence length results, which provide context for the downstream results (\S\ref{sec:results-representation-properties}). We then report downstream performance across pretraining and fine-tuning regimes, on both seen and unseen languages (\S\ref{sec:downstream}). In all result tables, boldface indicates a statistically significant improvement over the baseline under an approximate randomization test ($p < 0.05$).
%\sk{why not mention the fourth part about 4.1?}\sk{leaving a comment here but applies everywhere where a section is referenced. I typically use \S\ref{} to reference sections (as Yulia taught me haha) like \S\ref{sec:results-roman-wins}. It saves a lot of space}.\mz{changed format and added 4.1! Thanks for the tip Sachin and Yulia, I might pass it on someday haha.}%\mz{distinction deleted}\sk{i am hesistant to make this distinction, isn't romanized also phonology inspired, and IPA also doing script equalization.}.

\subsection{Properties of the Three Representations}
\label{sec:results-representation-properties}

% We report two properties of the three representations under their respective tokenizers. Both provide context for the results that follow.

\paragraph{Subword overlap between languages.}

\begin{figure}[t]
    \centering
    \includegraphics[width=\columnwidth]{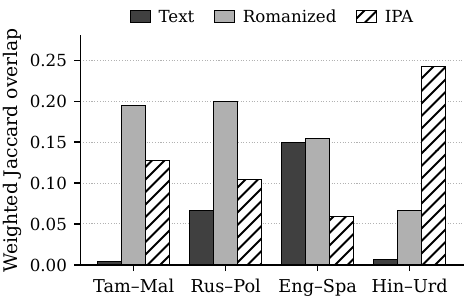}
\caption{Frequency-weighted subword overlap between languages within each pair under a shared 100K multilingual tokenizer, normalized for corpus size.}
    \label{fig:tokenizer-overlap}
\end{figure}

\begin{figure}[!t]
\centering
\includegraphics[width=\columnwidth]{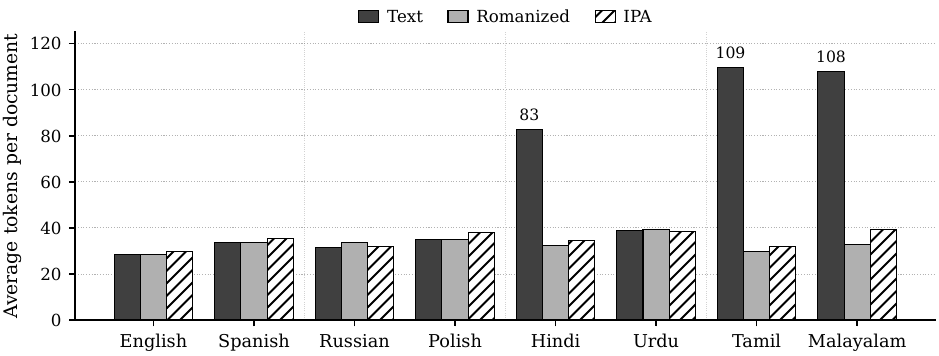}
\caption{Average sequence length (tokens per document) by language and representation on the FLORES parallel corpus.}
\label{fig:seq-length}
\end{figure}

\begin{table*}[!t]
\centering
% \scriptsize
\resizebox{\textwidth}{!}{
\setlength{\tabcolsep}{3pt}
\begin{tabular}{ll ccccccccc ccccccccc}
\toprule
& & \multicolumn{9}{c}{\textbf{XNLI}} & \multicolumn{9}{c}{\textbf{MASSIVE}} \\
\cmidrule(lr){3-11} \cmidrule(lr){12-20}
\textbf{Size} & \textbf{Rep.}
& \textbf{EN} & \textbf{ES} & \textbf{HI} & \textbf{UR} & \textbf{RU} & \textbf{PL} & \textbf{TA} & \textbf{ML} & \textbf{Macro}
& \textbf{EN} & \textbf{ES} & \textbf{HI} & \textbf{UR} & \textbf{RU} & \textbf{PL} & \textbf{TA} & \textbf{ML} & \textbf{Macro} \\
\midrule
\multirow{4}{*}{Small}
& Text       & 75.88 & 73.78 & 64.48 & 60.78 & 70.89 & 79.12 & 69.85 & 69.84 & 69.36
              & 75.56 & 72.62 & 71.13 & 63.32 & 73.10 & 71.02 & 65.58 & 66.16 & 69.81 \\
& IPA        & 76.22 & 73.58 & 68.07 & 63.73 & 69.58 & 78.40 & \textbf{71.11} & 70.84 & 70.45
              & 76.87 & 74.72 & 74.08 & 71.63 & 71.75 & 72.68 & \textbf{70.17} & 69.21 & 72.64 \\
& Romanized  & \textbf{78.99} & \textbf{76.25} & \textbf{69.65} & \textbf{64.98} & \textbf{73.86} & \textbf{81.81} & \textbf{71.80} & \textbf{72.71} & \textbf{72.61}
              & \textbf{79.90} & \textbf{77.80} & \textbf{75.70} & \textbf{73.70} & \textbf{77.80} & \textbf{77.40} & \textbf{69.80} & \textbf{71.90} & \textbf{75.50} \\
& Text$\rightarrow$Rom. & 74.67 & 71.72 & 56.47 & 59.11 & 62.77 & 56.71 & 62.23 & 64.07 & 63.47
              & 66.17 & 62.00 & 33.46 & 29.32 & 41.19 & 60.29 & 32.15 & 37.42 & 45.25 \\
\midrule
\multirow{4}{*}{Medium}
& Text       & 81.98 & 79.08 & 70.97 & 68.75 & 70.99 & 79.18 & \textbf{76.87} & 75.76 & 74.91
              & 81.98 & 80.31 & \textbf{80.60} & 75.31 & 80.95 & 78.49 & \textbf{77.64} & \textbf{80.05} & \textbf{79.42} \\
& IPA        & 81.09 & 78.96 & 73.08 & 68.55 & 76.38 & \textbf{84.30} & \textbf{75.95} & 76.28 & 75.76
              & \textbf{82.77} & 79.19 & \textbf{80.97} & \textbf{78.68} & 80.10 & \textbf{80.65} & 76.57 & \textbf{79.15} & \textbf{79.76} \\
& Romanized  & \textbf{83.18} & \textbf{80.49} & \textbf{74.20} & \textbf{70.37} & \textbf{77.82} & \textbf{84.20} & \textbf{76.86} & \textbf{77.55} & \textbf{77.21}
              & \textbf{83.04} & \textbf{81.40} & \textbf{80.06} & 77.36 & \textbf{82.54} & \textbf{81.45} & \textbf{76.84} & 78.21 & \textbf{80.11} \\
& Text$\rightarrow$Rom. & \textbf{82.63} & \textbf{80.06} & 62.95 & 58.76 & 67.15 & 59.10 & 65.09 & 65.15 & 67.61
              & 74.51 & 71.18 & 51.48 & 47.34 & 56.99 & 71.12 & 44.86 & 50.61 & 58.51 \\
\midrule
\multirow{4}{*}{Large}
& Text       & 84.05 & 81.40 & 76.55 & 71.44 & 79.88 & \textbf{88.30} & 78.40 & 77.37 & 79.67
              & 83.99 & 82.21 & 84.67 & 79.89 & 83.59 & 82.95 & \textbf{82.28} & \textbf{83.96} & 82.94 \\
& IPA        & \textbf{86.61} & \textbf{83.65} & \textbf{78.84} & 72.08 & 80.26 & \textbf{89.10} & \textbf{79.72} & \textbf{78.98} & \textbf{81.16}
              & \textbf{86.75} & 83.96 & \textbf{85.81} & \textbf{82.48} & 84.80 & \textbf{84.57} & \textbf{82.55} & \textbf{84.67} & \textbf{84.45} \\
& Romanized  & \textbf{86.69} & \textbf{84.13} & \textbf{79.00} & \textbf{73.23} & \textbf{81.68} & \textbf{88.70} & \textbf{79.80} & \textbf{79.56} & \textbf{81.60}
              & \textbf{86.42} & \textbf{85.88} & \textbf{86.28} & \textbf{82.72} & \textbf{86.28} & \textbf{85.27} & \textbf{82.28} & \textbf{84.30} & \textbf{84.93} \\
& Text$\rightarrow$Rom. & 84.33 & 81.36 & 65.58 & 62.75 & 70.13 & 62.48 & 70.88 & 65.50 & 70.38
              & 79.05 & 75.35 & 57.13 & 53.19 & 61.63 & 76.09 & 52.49 & 57.16 & 64.01 \\
\bottomrule
\end{tabular}
}
\caption{Fine-tuning F1 (\%) on the eight pretraining languages. Bold marks the best representation and statistical ties per benchmark, scale, and language. \textit{Text$\rightarrow$Rom.} denotes text-pretrained models fine-tuned on romanized data.}
\label{tab:seen-finetuning-xnli-massive}
\end{table*}

\autoref{fig:tokenizer-overlap} reports frequency-weighted subword overlap between the two languages in each typological pair under the shared tokenizer. Under orthographic text, meaningful overlap appears only for the Latin-script English--Spanish pair; the three different-script pairs sit at or near zero. Romanization lifts two of those three pairs (Russian--Polish and Tamil--Malayalam) into a range comparable to English--Spanish, but does not close the gap for Hindi--Urdu. IPA inverts this pattern: it produces its highest overlap on Hindi--Urdu, where phonology is nearly identical despite disjoint orthographies, but lags romanization on every other pair. Romanization and IPA are therefore not interchangeable. Prior work has reached mixed conclusions on the role of overlap in cross-lingual transfer \citep{kk2020crosslingual, limisiewicz2023tokenization, meyer2024systematic}. We also treat overlap as one channel among several. %, consistent with these findings.

\paragraph{Sequence length.}

\autoref{fig:seq-length} reports average tokens per document on the FLORES parallel corpus, where the same content is expressed across all eight languages \citep{nllb2022nollanguageleftbehind}. % and per-token information density is directly comparable. 
Compared to the other five, Hindi, Tamil, and Malayalam tokens inflate under text by factors of roughly three times their romanized counterparts. IPA and romanization compress them back into the same range as the others, consistent with \citet{anonymous2026phonemes} who report similar compression gains under IPA tokenization. With text, the inflated languages require roughly three times the inference latency and API costs compared with Latin- and Cyrillic-script peers, and romanization and IPA close this gap.%\sk{could cite that IPA ACL2026 paper here.}%\mz{added citation}

\subsection{Downstream Performance}
\label{sec:downstream}

\begin{table}[t]
\centering
% \scriptsize
\resizebox{\columnwidth}{!}{
\setlength{\tabcolsep}{4pt}
\begin{tabular}{lllccccc}
\toprule
\textbf{Setting} & \textbf{Size} & \textbf{Rep.}
& \textbf{AR} & \textbf{BN} & \textbf{EL} & \textbf{FR}
& \textbf{Macro} \\
\midrule
\multirow{9}{*}{ZS}
& \multirow{3}{*}{Small}
& Text       & 6.84  & 0.94  & 6.03  & \textbf{21.97} & 8.95 \\
& & IPA      & 5.66  & 10.25 & \textbf{10.74} & 6.96  & 8.40 \\
& & Romanized& \textbf{8.57}  & \textbf{12.87} & \textbf{10.43} & 27.68 & \textbf{14.89} \\
\cmidrule(lr){2-8}
& \multirow{3}{*}{Medium}
& Text       & \textbf{11.06} & 2.38  & 7.10  & 31.22 & 12.94 \\
& & IPA      & 7.87  & 19.82 & \textbf{13.50} & 13.98 & 13.79 \\
& & Romanized& 9.34  & \textbf{22.47} & \textbf{13.37} & \textbf{39.29} & \textbf{21.12} \\
\cmidrule(lr){2-8}
& \multirow{3}{*}{Large}
& Text       & \textbf{15.67} & 3.56 & 11.60  & 36.11 & 16.73 \\
& & IPA      & 13.01 & 25.66 & \textbf{19.20} & 21.49 & 19.84 \\
& & Romanized& 13.27 & \textbf{31.24} & \textbf{19.33} & \textbf{44.35} & \textbf{27.05} \\
\midrule
\multirow{12}{*}{FT}
& \multirow{4}{*}{Small}
& Text       & 34.63 & 30.66 & 26.59 & 47.35 & 34.81 \\
& & IPA      & 35.93 & \textbf{48.72} & 49.15 & 45.73 & 44.88 \\
& & Romanized& \textbf{41.14} & \textbf{48.84} & \textbf{52.24} & \textbf{54.41} & \textbf{49.16} \\
& & Text$\rightarrow$Rom. & 31.78 & 36.63 & 40.10 & 39.17 & 36.92 \\
\cmidrule(lr){2-8}
& \multirow{4}{*}{Medium}
& Text       & 51.87 & 51.77 & 54.05 & 62.15& 54.96 \\
& & IPA      & 53.68 & 59.76 & \textbf{64.94} & 61.21 & 59.90 \\
& & Romanized& \textbf{55.28} & \textbf{61.13} & \textbf{65.78} & \textbf{65.87} & \textbf{62.02} \\
& & Text$\rightarrow$Rom. & 45.24 & 49.68 & 53.44 & 53.76 & 50.53 \\
\cmidrule(lr){2-8}
& \multirow{4}{*}{Large}
& Text       & 57.53 & 59.15 & 61.47 & 67.52 & 61.42 \\
& & IPA      & 61.87 & 68.36 & \textbf{70.98} & 68.93 & 67.53 \\
& & Romanized& \textbf{64.69} & \textbf{70.11} & \textbf{71.25} & \textbf{75.62} & \textbf{70.42} \\
& & Text$\rightarrow$Rom. & 49.58 & 58.58 & 61.39 & 60.30 & 57.46 \\
\bottomrule
\end{tabular}
}
\caption{Cross-lingual transfer to unseen languages on MASSIVE, F1 (\%). %\sk{i have correct it at other places, but please just use \% instead of x100}. \mz{changed all captions} 
\textit{Zero-Shot} evaluates pretrained models directly; \textit{Unseen FT} fine-tunes on the unseen language. Bold marks the best representation and its statistical ties per setting, scale, and language.}
\label{tab:unseen-massive}
\end{table}

\begin{figure*}[!t]
\centering
\includegraphics[width=\textwidth]{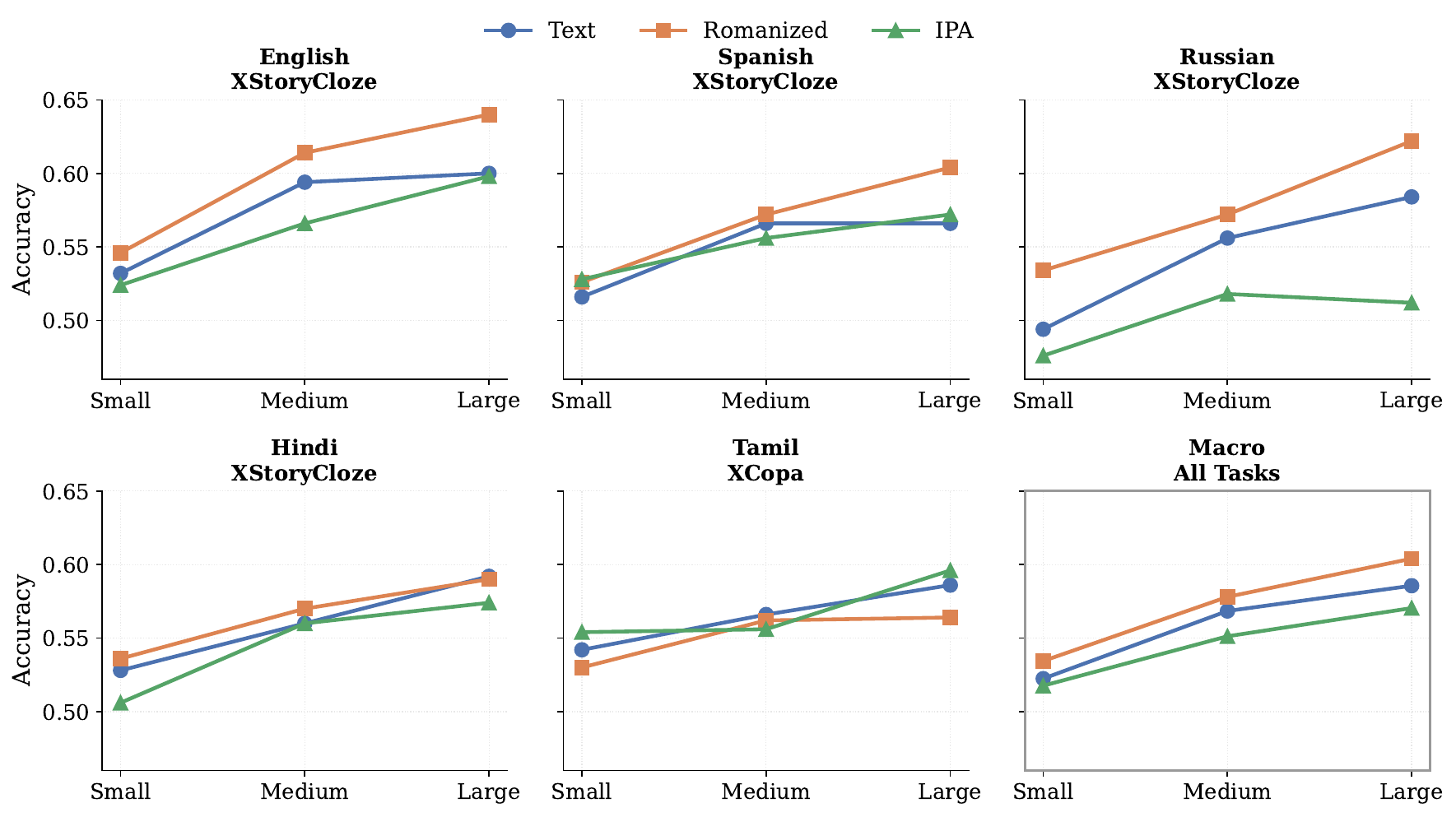}
\caption{Zero-shot accuracy on XStoryCloze and XCOPA across scales for the three representations. Each panel reports one evaluation language.}
\label{fig:icl-trends}
\end{figure*}

\begin{table}[!t]
\centering
\scriptsize
\setlength{\tabcolsep}{3pt}
\resizebox{\columnwidth}{!}{%
\begin{tabular}{llccccccc}
\toprule
\textbf{Size} & \textbf{Rep.}
& \textbf{EN} & \textbf{ES} & \textbf{HI} & \textbf{RU} & \textbf{TA} & \textbf{UR}
& \textbf{Macro} \\
\midrule
\multirow{3}{*}{Small}
& Text       & 17.47 & 15.70 & 12.85 & 10.91 & 3.48  & 21.47 & 13.65 \\
& IPA        & 11.33 & 13.41 & 16.66 & 9.15  & 14.22 & 18.76 & 12.26 \\
& Romanized  & \textbf{20.49} & \textbf{17.71} & \textbf{23.97} & \textbf{15.93} & \textbf{15.89} & \textbf{23.23} & \textbf{19.54} \\
& Text$\rightarrow$Rom. & 20.12 & 17.13 & 12.31 & 8.91  & 9.09  & 20.09 & 14.63 \\
\midrule
\multirow{3}{*}{Medium}
& Text       & 22.10 & 20.49 & 22.15 & 16.54 & 12.23 & 26.82 & 20.06 \\
& IPA        & 21.00 & 18.63 & \textbf{29.95} & 19.74 & \textbf{20.99} & 29.75 & 23.34 \\
& Romanized  & \textbf{24.04} & \textbf{20.79} & 28.17 & \textbf{19.75} & 20.91 & \textbf{30.83} & \textbf{24.08} \\
& Text$\rightarrow$Rom. & 26.51 & 22.21 & 21.91 & 13.02 & 10.02 & 18.55 & 18.73 \\
\midrule
\multirow{3}{*}{Large}
& Text       & 23.51 & 20.79 & 22.55 & 18.17 & 17.36 & 29.36 & 21.96 \\
& IPA        & 23.96 & 21.15 & \textbf{29.30} & 20.31 & 22.32 & \textbf{32.69} & 24.95 \\
& Romanized  & \textbf{25.52} & \textbf{21.23} & 28.45 & \textbf{22.17} & \textbf{22.68} & 31.95 & \textbf{25.33} \\
& Text$\rightarrow$Rom. & 27.18 & 22.83 & 20.45 & 15.56 & 13.38 & 22.42 & 20.30 \\
\bottomrule
\end{tabular}}
\caption{XL-Sum summarization, ROUGE-L F1 (\%). Bold marks the best representation per scale and language.}
\label{tab:xlsum-rougel}
\end{table}

\begin{table}[t]
\centering
\scriptsize
\setlength{\tabcolsep}{4pt}
\begin{tabular}{llccccc}
\toprule
\textbf{Size} & \textbf{Rep.}
& \textbf{AR} & \textbf{BN} & \textbf{EL} & \textbf{FR}
& \textbf{Macro} \\
\midrule
\multirow{4}{*}{Small}
& Text       & 61.88 & 65.37 & \textbf{66.02} & 66.01 & 64.82 \\
& IPA        & 61.70 & 65.31 & \textbf{65.74} & 64.64 & 64.35 \\
& Romanized  & \textbf{63.61} & \textbf{67.32} & \textbf{66.66} & \textbf{68.25} & \textbf{66.46} \\
& Text$\rightarrow$Rom. & \textbf{62.86} & 65.25 & \textbf{65.92} & 66.98 & 65.25 \\
\cmidrule(lr){1-7}
\multirow{4}{*}{Medium}
& Text       & 63.31 & 64.16 & 66.38 & 69.20 & 65.76 \\
& IPA        & 63.85 & 68.33 & 68.24 & 67.00 & 66.86 \\
& Romanized  & \textbf{68.89} & \textbf{72.56} & \textbf{71.23} & \textbf{75.07} & \textbf{71.94} \\
& Text$\rightarrow$Rom. & 64.72 & 68.11 & 68.47 & 70.61 & 67.98 \\
\cmidrule(lr){1-7}
\multirow{4}{*}{Large}
& Text       & 65.64 & 66.16 & 68.43 & 72.33 & 68.14 \\
& IPA        & 65.63 & 71.54 & 70.52 & 69.14 & 69.21 \\
& Romanized  & \textbf{71.27} & \textbf{75.63} & \textbf{72.74} & \textbf{78.27} & \textbf{74.48} \\
& Text$\rightarrow$Rom. & 66.59 & 69.06 & 69.68 & 71.78 & 69.28 \\
\bottomrule
\end{tabular}
\caption{Cross-lingual transfer to unseen languages on XNLI, F1 (\%). Models are fine-tuned on the unseen language's training set. Bold marks the best representation and its statistical ties per scale and language.}
\label{tab:unseen-xnli}
\end{table}

\begin{table}[t]
\centering
\small
\resizebox{\columnwidth}{!}{%
\begin{tabular}{llcccc}
\toprule
\textbf{Size} & \textbf{Rep.} & \textbf{AR} & \textbf{BN} & \textbf{FR} & \textbf{Macro} \\
\midrule
\multirow{4}{*}{Small}
& Text            & 7.21  & 2.64  & 16.40 & 8.75  \\
& IPA             & 4.39  & 6.64  & 13.29 & 8.11  \\
& Romanized       & \textbf{10.72} & \textbf{9.08} & \textbf{17.79} & \textbf{12.53} \\
& Text$\rightarrow$Rom. & 10.06 & 6.48 & 15.83 & 10.79 \\
\midrule
\multirow{4}{*}{Medium}
& Text            & 10.72 & 5.32  & 18.37 & 11.14 \\
& IPA             & 9.25  & 9.58  & 16.90 & 11.91 \\
& Romanized       & \textbf{12.45} & \textbf{11.80} & \textbf{19.57} & \textbf{14.61} \\
& Text$\rightarrow$Rom. & 11.84 & 9.54 & 17.62 & 13.00 \\
\midrule
\multirow{4}{*}{Large}
& Text            & 12.58 & 7.31  & 17.38 & 12.42 \\
& IPA             & 11.29 & 11.76 & 20.03 & 14.36 \\
& Romanized       & \textbf{15.36} & \textbf{12.43} & \textbf{21.43} & \textbf{16.41} \\
& Text$\rightarrow$Rom. & 12.76 & 8.99 & 18.20 & 13.25 \\
\bottomrule
\end{tabular}}
\caption{XL-Sum cross-lingual transfer to unseen languages, ROUGE-L F1 (\%). Bold marks the best representation per scale and language.}
\label{tab:unseen-xlsum}
\end{table}

% \sk{suggestion: list/ref all the tables/figures here first. Each of the subsequent titled paragraphs are just restating the finding in already mentioned in this paragraph, unless there's truly interesting things to say, it does not need to be repeated. We don't need separate titled paragraphs for incontext and generation etc.}\mz{cut down content and adjusted order between ICL and FT.}
\paragraph{Multilingual romanized pretraining is the strongest.} Pretraining on romanized text produces the strongest cross-lingual transfer across every regime we evaluate: zero- and few-shot in-context learning on XStoryCloze and XCOPA (\autoref{fig:icl-trends}; full k-shot results in \autoref{app:ICL},  \autoref{tab:scale-kshot-results}; the same ordering holds at intermediate pretraining checkpoints, ), 
supervised fine-tuning on NLI and MASSIVE (\autoref{tab:seen-finetuning-xnli-massive}) and XL-Sum summarization (\autoref{tab:xlsum-rougel}), and transfer to unseen languages on MASSIVE (\autoref{tab:unseen-massive}) and XNLI (\autoref{tab:unseen-xnli}). The effect holds at all three model scales.
% \mz{deleted}\sk{this is not adding new info right? could delete}. 
Under prompted setups, the gap to text is smaller than under fine-tuning, but the ranking is preserved. %We report complete zero shot and in-context learning results with $k \in \{1, 2, 4\}$ in-context examples in \autoref{app:ICL}.

IPA improves over text in most settings we evaluate. This is consistent with \citet{jung2026happiness}, who find that IPA improves over orthographic text in encoder-only models, especially in unseen languages, and contrasts with prior works \citep{goriely2024babble, bunzeck2024graphemes}, %\sk{please double check that these actually contrast. i think jung2024 have mixed results.} \mz{modified}, 
who report phoneme-trained language models slightly underperform their text-trained counterparts. Outside Hindi and Urdu, which share little orthography but are nearly identical phonologically, IPA lags romanization, especially on unseen languages. 

To our knowledge, no prior work has directly compared phonemic and romanized pretraining for autoregressive LMs; our results provide the first such evidence and show that the two are not interchangeable. %. On unseen MASSIVE, IPA lags romanized substantially under both zero-shot and fine-tuning.%\sk{this is good but we should mention explicitly that is does improve over text is most cases, as opposed to claims made by some prior work. Also we can say that no prior work has compared romanized and IPA and we provide that evidence.} \mz{added one prior work as contrast and a contribution statement.} 
%On unseen XNLI, IPA does not improve over text, with the macro gap remaining within half a point at every scale. 
We attribute this to two factors: %\sk{i don't understand what open-endedness means here. IPA is also a small set of characters, not that much larger than Latin alphabet. Also a major reason we could say is that IPA itself might not be good quality of many languages.}. 
First, Phonemizer's %\sk{we haven't mentioned espeak before, could just call it phonemizer. Also discuss why other better phonemization tools are slow at scale. we could say that more research is needed to make definite claims.} \mz{added in the footnote in section 3.3} phonemization 
quality varies across different languages. The phonemization is noisier, especially for low-resource languages \citep{Bernard2021, goriely2024babble}. Second, the phonemized output of an unseen language may also include symbols not produced by the eight pretraining languages, which would compound the problem.  
%XL-Sum (Table~\ref{tab:xlsum-rougel}) extends the finding to summarization, where the model must produce surface forms rather than select among label candidates. Romanized pretraining achieves the highest macro ROUGE-L at every scale and wins on at least four of the six languages at every scale.

% \sk{so far text->roman results haven't been discussed at all. I would mention that first followed by your GPT2 experiment. } \mz{they are discussed in the next subsection. maybe we should swap their orders so multilingual text->roman is discussed first?}\sk{yes, otherwise it seems a bit jarring to jump here. A way to frame this would: we first claim that text->roman is worse than text->text surprisingly different from findings of prior work. We then highlight the crucial difference: prior work only tested with primarily English-only pretrained models. So we did controlled study. We finetuning a model trained only on English (Gpt2) and then the gains come back. There's two implications of this: all new pretrained models are multilingual, so the findings of previous works are invalidated, but all is not lose, future work should focus on romanized pretraining at such scales.}\sk{also same before, we don't need titled paragraphs for every finding.} \mz{swapped order and modified contents.}

\begin{table*}[!t]
\centering
\scriptsize
\setlength{\tabcolsep}{3pt}
\begin{tabular}{ll ccccccccc ccccccc}
\toprule
& & \multicolumn{9}{c}{\textbf{MASSIVE}} & \multicolumn{7}{c}{\textbf{XL-Sum}} \\
\cmidrule(lr){3-11} \cmidrule(lr){12-18}
\textbf{Backbone} & \textbf{Rep.}
& \textbf{EN} & \textbf{ES} & \textbf{PL} & \textbf{RU} & \textbf{HI} & \textbf{UR} & \textbf{ML} & \textbf{TA} & \textbf{Macro}
& \textbf{EN} & \textbf{ES} & \textbf{RU} & \textbf{HI} & \textbf{UR} & \textbf{TA} & \textbf{Macro} \\
\midrule
\multirow{2}{*}{GPT-2}
& Text & \textbf{54.0} & \textbf{30.1} & \textbf{28.4} & 16.6 & 9.0 & 9.3 & 8.7 & 5.2 & 20.2
              & 21.67 & 14.18 & 3.35 & 0.47 & 0.53 & 0.00 & 6.66 \\
& Romanized & 44.0 & 25.1 & 26.9 & \textbf{20.2} & \textbf{20.5} & \textbf{17.1} & \textbf{23.5} & \textbf{20.3} & \textbf{24.7}
              & \textbf{21.94} & \textbf{14.57} & \textbf{8.46} & \textbf{10.52} & \textbf{13.84} & \textbf{4.53} & \textbf{12.31} \\
\midrule
\multirow{2}{*}{GPT-2 Medium}
& Text & \textbf{69.6} & \textbf{47.3} & 43.9 & 22.9 & 14.6 & 12.8 & 10.0 & 11.0 & 29.0
              & \textbf{24.83} & 14.25 & 3.24 & 0.43 & 0.52 & 0.00 & 7.18 \\
& Romanized & 66.1 & 43.3 & \textbf{45.3} & \textbf{39.0} & \textbf{42.7} & \textbf{34.0} & \textbf{40.2} & \textbf{35.5} & \textbf{43.3}
              & \textbf{24.85} & \textbf{14.36} & \textbf{9.12} & \textbf{10.64} & \textbf{14.27} & \textbf{4.88} & \textbf{13.03} \\
\midrule
\multirow{2}{*}{GPT-2 Large}
& Text & \textbf{80.0} & \textbf{64.9} & 61.5 & 40.0 & 32.7 & 33.1 & 26.0 & 21.6 & 45.0
              & 26.24 & 14.46 & 3.38 & 0.45 & 0.39 & 0.00 & 7.49 \\
& Romanized & 77.9 & 61.6 & \textbf{62.2} & \textbf{59.1} & \textbf{56.3} & \textbf{53.7} & \textbf{60.3} & \textbf{56.0} & \textbf{60.9}
              & \textbf{26.37} & \textbf{15.92} & \textbf{8.79} & \textbf{10.87} & \textbf{14.26} & \textbf{4.24} & \textbf{13.42} \\
\bottomrule
\end{tabular}
\caption{Fine-tuning English-pretrained GPT-2 checkpoints on MASSIVE (F1, \%) and XL-Sum (ROUGE-L, \%) under the Text and Romanized conditions. Bold marks the better representation per backbone, benchmark, and language. XL-Sum does not include ML.}
\label{tab:hf_gpt2_combined}
\end{table*}

\paragraph{Romanized finetuning a text-pretrained model degrades performance on seen languages.}%\sk{for saving space, i am thinking 4.2, 4.3, 4.4 and 4.5 could paragraphs under a new subsection 4.2 (which could be about downstream performance).}  \mz{changed into paragraphs} 
Given the strong results of romanization, a natural intervention is to finetune a text-pretrained model on romanized data, hoping to recover cross-lingual transfer benefits of romanization without paying the cost of romanized pretraining. \citet{husain2024romansetu} report that such an intervention helps when adapting Llama~2 to non-Latin-script languages, and it is the standard recipe in prior romanization-for-transfer work. We apply this intervention to our multilingual text-pretrained checkpoint. Reported as Text$\rightarrow$Rom in Tables~\ref{tab:seen-finetuning-xnli-massive} and \ref{tab:xlsum-rougel}, we find that, contrary to expectation, it produces large regressions on all pretraining languages across all benchmarks. On unseen languages where the pretrained model lacks script coverage (Greek and Bengali) in the same sense as an English-only model would, the intervention does help, though by much smaller margins than romanized pretraining.
% Tables~\ref{tab:seen-finetuning-xnli-massive} and \ref{tab:xlsum-rougel} report the Text$\rightarrow$Rom condition alongside the from-scratch baselines on the eight pretraining languages. On XNLI, Text$\rightarrow$Rom regresses several macro F1 points behind the matched Text$\rightarrow$Text baseline at every scale, and the gap does not close as the model grows. On MASSIVE the regression is far more severe and concentrates on the non-Latin-script languages, which lose 30 or more accuracy points at Small. XL-Sum shows the same direction in the generative setting, and ICL results on XStoryCloze (Appendix Table~\ref{tab:scale-kshot-results}) corroborate the picture on Russian and Hindi.\sk{this whole paragraph could be skipped}
%On unseen languages (Tables~\ref{tab:unseen-xnli} and \ref{tab:unseen-xlsum}), Text$\rightarrow$Rom outperforms the Text$\rightarrow$Text baseline at every scale and on every individual language. %\sk{gpt-2 finding hasn't been described before}. \mz{deleted}
Unseen MASSIVE (\autoref{tab:unseen-massive}) is a partial exception: the benefit holds at the smallest scale but disappears at Medium and Large. We attribute this to MASSIVE being more sensitive to surface-form noise, since intent classification depends on recognizing short trigger phrases that romanization can distort.

\paragraph{Romanized finetuning of a \textit{monolingual} text-pretrained model improves transfer} 
The previous result contradicts prior reports that Text$\rightarrow$Rom fine-tuning helps cross-lingual transfer. To further understand the cause of these regressions, we mirror the prior work setup: fine-tuning an English-only pretrained model \citep{radford2019gpt2} on romanized multilingual downstream data. As shown in~\autoref{tab:hf_gpt2_combined}, we do recover the improvements reported by prior work, but improvements achieved through this fine-tuning-time intervention fall well short of what pretraining-time romanization delivers, as reported in previous sections. %}\sk{this has already been said in the previous section. We can just say: to understand the cause of regressions in previous section, we finetuned english only model mirroring prior work, and do indeed recover performance improvement, but the gap is smaller.} \mz{modified. }
% Romanized fine-tuning improves macro performance at every scale on both benchmarks (Table~\ref{tab:hf_gpt2_combined}). The MASSIVE gap grows monotonically with scale, reaching roughly three times the smallest-model benefit; the XL-Sum gap is already large at the smallest scale and stable across scales, since the text baseline produces near-zero ROUGE-L on non-Latin-script languages and romanization recovers most of the lost capability.
The gains concentrate in languages with absent or rare scripts. Hindi, Urdu, Tamil, and Malayalam gain 20--30 F1 on MASSIVE at larger scales and move from near-zero to 10--15 on XL-Sum. Latin-scripted languages show small, sometimes negative changes. %; Russian sits between the two regimes.

These findings qualify \citet{husain2024romansetu}'s results: %by showing the conditions under which Text$\rightarrow$Rom. 
finetuning helps when the base model lacks coverage of the target script, and degrades performance when such coverage already exists. As most modern open-source and open-weight models are multilingual, covering a wide array of scripts, romanized finetuning will likely lead to uneven results. This asymmetry motivates future work on romanized multilingual pretraining at scale.

%}\sk{i would try to summarize this whole paragraph in 1-2 sentences max, and talk about future work like you did. much of this has been covered in the previous subsection.}\mz{modified}

% \paragraph{Script-equalization is the dominant driver of cross-lingual transfer}
\noindent \\
Taken together, these results produce a consistent ranking: romanizing at pretraining is best, while romanizing at fine-tuning helps when pretraining lacked script coverage and hurts when it established it. This echoes \citet{xhelili2024breaking} and \citet{liu2025transliterations}, who find that transliteration's benefits depend on exposing lexical overlap rather than on the operation itself: the same operation supplies overlap when the model lacks script-specific representations and disrupts it when those representations already exist. Characterizing how this plays out in the model's internal representations is a natural next step that we leave to future work.
%\sk{the comments about geometry are interesting but we have no experiments to back them up. Without those comments, this subsection has not much to say other than repeating the previous comments. I do get the point of transliteration and lexical overlap now. I feel like once could end the previous three subsections (after converting to paragraphs) with a new untitled paragraph where you summarize this subsection into 2-3 sentences, and discusing geometrical analysis as potential future work / or cite if there's such work already.}\mz{modified}

\section{Conclusion}
We study input representation as a controlled variable in autoregressive multilingual pretraining, comparing orthographic text, IPA, and Uroman romanization under matched conditions across three scales and eight languages. Romanized pretraining yields the strongest cross-lingual transfer on both seen and unseen languages. IPA improves over text in most settings but only matches romanization on Hindi--Urdu, where phonology is shared, and orthography is disjoint. The widely used recipe of fine-tuning a text-pretrained model on romanized data regresses performance on languages the model already covers, and helps only when the base model lacks script coverage. We recommend that future multilingual pretraining treat input representation as a deliberate design decision alongside data mixture and tokenizer, and adopt romanization when transfer to unseen scripts is a priority.

\section{Limitations}
\paragraph{Scale.}
Due to our limited compute budget, we pretrained models up to approximately 1B parameters (387M non-embedding), substantially smaller than contemporary multilingual language models, which typically range from 7B to over 100B parameters \citep{aryabumi2024aya23, grattafiori2024llama3, qwen2024qwen25, yang2025qwen3}. Future work may extend this work to larger scale settings.

\paragraph{Language coverage.}
We selected language pairs intended to highlight differences in orthography alongside varying degrees of phonetic overlap: English and Spanish, Russian and Polish, Hindi and Urdu, and Tamil and Malayalam. These languages primarily represent alphabetic or abugida-based writing systems. Many other classes of writing systems were not represented in this work, including logographic systems such as Chinese. Expanding this analysis to additional languages and orthographic systems remains an important direction for future work. %, particularly to better understand how phonetic representations interact with fundamentally different writing systems and linguistic structures.
Future work should also investigate how varying degrees of linguistic relatedness and shared inheritance, such as language family distance, phonological similarity, and the presence or absence of shared loanwords, affect the extent to which different representations improve transfer. %multilingual generalization.

\paragraph{Transcription Quality.}
% Transcription quality appears to come at the cost of processing speed (\S\ref{sec:g2p}). 
Phonemization is an active area of research. We elected to use the Phonemizer library because it provided a practical balance between transcription quality and throughput at the scale required for our corpus. Although we performed small-scale comparisons of different G2P tools, a more comprehensive evaluation of the computational cost introduced by transcription during both training and inference is needed. Improving phonemization could change our reported trends as it pertains to IPA. % remains an important direction for future work.

Additionally, the authors of \citet{espeak_languages}, the backend framework powering Phonemizer, note that support for several languages remains incomplete or insufficient, which may affect transcription consistency and overall model quality for low-resource or under-supported languages. During preprocessing, we also observed limitations in using standard Unicode representations for IPA tokenization, particularly with regard to handling compound symbols and diacritic composition, which led to stripping of diacritics and stress markers (\S\ref{sec:g2p}). During our analysis, we found evidence that Unicode may not be the best representation for phonemes for this kind of work; however, exploring alternative tokenization or encoding strategies for IPA representations is, therefore, left to future work. %\aj{I'm unsure if I need this section, it's relevant to the work I did for interpretability and the findings from there, but I don't think we directly discuss it anywhere in this paper aside from brief mentions}

% Our IPA normalization process was similarly incomplete. While several diacritics were removed to reduce representational complexity, certain markers, such as syllable boundaries (.), were preserved by accident. Further investigation is needed to determine whether retaining or removing these markers meaningfully impacts downstream model performance. 
Finally, our approach for IPA stripping and preserving numerical information during transcription (\S\ref{sec:g2p}) relied on a relatively simple heuristic-based system. More sophisticated approaches may improve robustness.

\paragraph{Converting IPA and romanized representations back to text.}
To be useful in generative settings, a language modeling-based system must operate in orthographic text, as a typical user might not be able to read romanized text or IPA. 
%Both romanization and phonemization are lossy operations. Hence, transcribing back to text might lead to ambiguities. 
There has been prior research exploring phoneme-to-grapheme (P2G) conversion \cite{lauc2024polyipamultilingualphonemetographeme, ma2025llmbasedphonemetographemephonemebasedspeech, masumura_phoneme_grapheme_2020}. However, we did not evaluate the performance or practicality of these approaches within our pipeline. %We did not quantify the computational tradeoff between model inference and the cost of repeatedly converting between orthographic and phonetic representations. 
% A more thorough analysis of their quality and costs is left for future work.

Existing work on P2G also presents several limitations. Both romanization and phonemization are lossy operations. For example, \citet{hermjakob-etal-2018-box} note that the transformation process is not fully reversible, potentially resulting in information loss during reconstruction. Many of these approaches are language-specific \cite{shibli_automatic_2023, baruah_assamesebacktranslit_2024}, though some more generalized approaches have been proposed, such as \citet{sequiera_word-level_2014}. Evaluating the effectiveness, reversibility, and computational overhead of these systems remains an important direction for future work.

\bibliography{custom}

\appendix

\section{Tokenizer Overlap Computation Details}
\label{app:overlap-details}

We compute cross-lingual subword overlap as follows. Given the multilingual tokenizer and two languages $L_1$ and $L_2$, we tokenize each language's monolingual corpus separately and let $T_{L_i}$ denote the set of unique tokens observed in $L_i$. The simplest measure of shared vocabulary is the Jaccard index over these token sets:
\begin{equation}
    J(L_1, L_2) = \frac{|T_{L_1} \cap T_{L_2}|}{|T_{L_1} \cup T_{L_2}|}.
\end{equation}
Higher values indicate greater shared vocabulary and, by the argument in section~\ref{sec:tokenization}, greater potential for transfer through shared embeddings \citep{conneau2020}.

Raw Jaccard, however, is sensitive to several confounds: rare tokens contribute equally to common ones, punctuation and whitespace artifacts inflate apparent overlap, code-switched lines leak tokens across languages, and unequal corpus sizes bias the unique-token counts. We therefore complement raw Jaccard with an adjusted variant that applies four corrections. First, we filter for contamination by retaining only documents whose language tag matches the target language and removing code-switched or noisy lines (motivated by language-switch warnings observed during preprocessing). Second, we exclude punctuation- and whitespace-only tokens, which would otherwise dominate the intersection. Third, we replace set-based Jaccard with a frequency-aware weighted variant that weights each token by its actual usage in the two corpora:
\begin{equation}
J_w(L_1, L_2) =
\frac{\sum_t \min(c_{L_1}(t), c_{L_2}(t))}
     {\sum_t \max(c_{L_1}(t), c_{L_2}(t))},
\end{equation}
where $c_{L_i}(t)$ is the count of token $t$ in the corpus of $L_i$. This downweights tokens that are technically shared but rare in one language. Fourth, we control for corpus-size imbalance by sampling matched token budgets per language and averaging the resulting overlap over multiple random samples.

\section{Pretraining details}
\label{app:pretraining}

\paragraph{Training procedure.}
            
We use a dual-optimizer setup: AdamW ($\beta_1{=}0.8$, $\beta_2{=}0.95$) for embedding, head, and scalar parameters, and Muon (momentum${=}0.95$) for hidden-layer matrix parameters, with a linear cooldown learning-rate schedule (peak Muon learning rate as $0.025$). Each training step processes 524{,}288 tokens across 8 NVIDIA H100 GPUs. %\sk{how long do you train for? we could also include training/validation logs in the appendix.} \mz{added training time}
We train each model for 3 epochs over the training split of the pretraining corpus, and select the checkpoint with the lowest validation loss on the held-out split for downstream evaluation.%\aj{I moved this here, you can rework it if it doesn't match or fit.}

\begin{table}[t]
    \centering
    \small
    \begin{tabular}{lccc}
        \toprule
        \textbf{Representation} & \textbf{Small} & \textbf{Medium} & \textbf{Large} \\
        \midrule
        Text       & 1.365 & 1.043 & 0.868 \\
        IPA        & 1.143 & 0.899 & 0.818 \\
        Romanized  & 1.087 & 0.862 & 0.775 \\
        \bottomrule
    \end{tabular}
    \caption{Multilingual pretraining bits per character by representation across scales. Values are not directly comparable across representations.}
    \label{tab:bpc-multilingual}
\end{table}

\begin{table}[t]
    \centering
    \small
    \setlength{\tabcolsep}{3pt}
    \begin{tabular}{lcccc}
        \toprule
        \textbf{Rep.} & \textsc{eng}--\textsc{spa} & \textsc{hin}--\textsc{urd} & \textsc{rus}--\textsc{pol} & \textsc{tam}--\textsc{mal} \\
        \midrule
        IPA (stripped) & 0.989 & 0.897 & 0.893 & 0.872 \\
        Text (orig.)   & 0.994 & 0.703 & 0.997 & 1.044 \\
        Phonemes       & 0.992 & \textbf{0.650} & \textbf{0.786} & \textbf{0.712} \\
        Romanized      & \textbf{0.945} & 0.735 & 0.937 & 0.808 \\
        \bottomrule
    \end{tabular}
    \caption{Pretraining bits-per-character (BPC) across input representations and language pairs at the \textbf{Small} scale. Bold indicates the best representation per language pair. 
    Lower is better.}
    \label{tab:bpc-results-small}

\end{table}

\begin{table}[t]
    \centering
    \small
    \setlength{\tabcolsep}{3pt}
    \begin{tabular}{lcccc}
        \toprule
        \textbf{Rep.} & \textsc{eng}--\textsc{spa} & \textsc{hin}--\textsc{urd} & \textsc{rus}--\textsc{pol} & \textsc{tam}--\textsc{mal} \\
        \midrule
        IPA (stripped) & 0.838 & 0.765 & 0.723 & 0.759 \\
        Text (orig.)   & 0.857 & \textbf{0.509} & 0.823 & 0.784 \\
        Phonemes       & 0.828 & 0.556 & \textbf{0.650} & \textbf{0.618} \\
        Romanized      & \textbf{0.812} & 0.649 & 0.783 & 0.704 \\
        \bottomrule
    \end{tabular}
    \caption{Pretraining bits-per-character (BPC) across input representations and language pairs at the \textbf{Medium} scale. Bold indicates the best representation per language pair. Lower is better.}
    \label{tab:bpc-results-medium}
\end{table}

\paragraph{Bits per character.}
We report multilingual pretraining bits per character (BPC) by representation across model scales in \autoref{tab:bpc-multilingual}, and bilingual pretraining BPC at the per-pair level in \autoref{tab:bpc-results-small} and \autoref{tab:bpc-results-medium} for small and medium models, respectively. We caution that BPC values normalized by representation-specific characters are not directly comparable across representations with different vocabularies and surface forms, since the denominator differs across configurations. We report these numbers for completeness and as evidence that all configurations trained successfully; we do not draw cross-representation conclusions from them, and rely instead on downstream task performance (\S\ref{sec:downstream}) as the operational measure of pretraining quality.

\paragraph{Pretraining-corpus sequence lengths.}

\begin{figure}[t]
\centering
\includegraphics[width=\columnwidth]{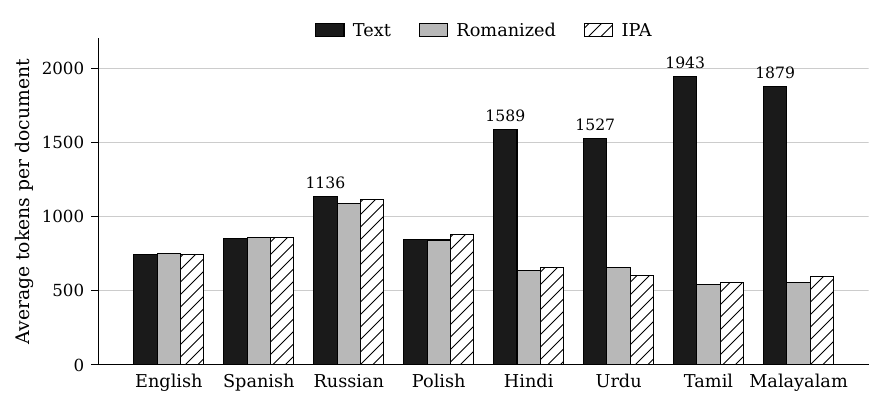}
\caption{Average sequence length (tokens per document) by language and representation on the pretraining corpus.}
\label{fig:seq-length-pretraining}
\end{figure}

\autoref{fig:seq-length-pretraining} reports average tokens per document on our pretraining corpus, FineWeb-2. % Document content differs across languages, but the within-language comparison across representations remains controlled.

Compared to the four Latin- and Cyrillic-script languages, Hindi, Urdu, Tamil, and Malayalam tokens inflate under text by factors of roughly 2.5--3.5 times their romanized counterparts, with the largest gaps for Tamil and Malayalam. IPA and romanization compress them back into the same range as the others. Under a fixed per-step token budget, text models therefore cover 2.5--3.5 times fewer documents per step in these four languages than IPA or romanized models do, and romanization and IPA close this gap during training.

\section{Fine-tuning Details}
\label{app:finetuning-details}

\subsection{Training Procedure}

For each \textit{(task, representation, scale)} cell, we perform a grid search over learning rate $\in \{5{\times}10^{-5}, 1{\times}10^{-4}, 2{\times}10^{-4}, 3{\times}10^{-4}\}$ and batch size $\in \{128, 256\}$, training for three epochs with linear warmup and decay. We select hyperparameters based on the lowest validation loss. For seen languages, we fine-tune jointly across all languages in a single multilingual run; for unseen languages, we fine-tune a separate model per language.

\subsection{Bilingual Downstream Results}
\label{app:bilingual-downstream}

\begin{table}[h]
\centering
\small
\setlength{\tabcolsep}{4pt}
\caption{Bilingual cross-lingual transfer F1 at the small model scale. Each cell reports the best F1 across transfer directions and hyperparameters. Bold marks the best representation per language pair.}
\label{tab:model-performance}
\begin{tabular}{llccc}
\toprule
\textbf{Dataset} & \textbf{Pair} & \textbf{Text} & \textbf{IPA} & \textbf{Rom.} \\
\midrule
\multirow{3}{*}{XNLI}
  & Eng--Spa & 0.85          & 0.81          & \textbf{0.86} \\
  & Hin--Urd & 0.34          & \textbf{0.64} & 0.51          \\
  & Rus--Pol & 0.35          & 0.40          & \textbf{0.43} \\
\midrule
IndicXNLI & Tam--Mal & 0.68 & 0.68 & \textbf{0.71} \\
\midrule
\multirow{4}{*}{MASSIVE}
  & Eng--Spa & \textbf{0.588} & 0.332          & 0.578          \\
  & Hin--Urd & 0.030          & \textbf{0.400} & 0.040          \\
  & Tam--Mal & 0.213          & 0.500          & \textbf{0.573} \\
  & Rus--Pol & 0.490          & 0.480          & \textbf{0.610} \\
\bottomrule
\end{tabular}
\end{table}
Table~\ref{tab:model-performance} reports cross-lingual transfer performance for bilingual models at a small scale. These results informed the design of our multilingual experiments presented in the main body.

Romanization or IPA outperforms text on seven of eight (dataset, pair) settings; text wins only on MASSIVE Eng--Spa, the one pair that already shares Latin script. The gaps are largest for related-language pairs with disjoint scripts, where text often collapses to near-zero F1 while IPA and romanization recover substantial transfer. Romanization is the most consistent winner across settings, while IPA's largest gains concentrate on Hin--Urd, the pair with the closest phonological overlap and the most disjoint scripts.

\section{In-Context Learning}
\label{app:ICL}

\begin{table*}[!t]
\centering
\small
\setlength{\tabcolsep}{4pt}
\renewcommand{\arraystretch}{1.05}
\begin{tabular}{llp{0.18\textwidth}cccccccccccc}
\toprule
& & & \multicolumn{4}{c}{\textbf{Small}} & \multicolumn{4}{c}{\textbf{Medium}} & \multicolumn{4}{c}{\textbf{Large-std}} \\
\cmidrule(lr){4-7} \cmidrule(lr){8-11} \cmidrule(lr){12-15}
\textbf{Task} & \textbf{Lang} & \textbf{Repr.} & $k{=}0$ & $k{=}1$ & $k{=}2$ & $k{=}4$ & $k{=}0$ & $k{=}1$ & $k{=}2$ & $k{=}4$ & $k{=}0$ & $k{=}1$ & $k{=}2$ & $k{=}4$ \\
\midrule
\multirow{3}{*}{XStoryCloze} & \multirow{3}{*}{en}
  & text      & .532 & .540 & \textbf{.552} & .524 & .594 & .596 & \textbf{.600} & .594 & .600 & .606 & .610 & \textbf{.612} \\
& & romanized & .546 & .550 & \textbf{.552} & .522 & .614 & .602 & \textbf{.618} & .604 & \textbf{.640} & .612 & .622 & .622 \\
& & ipa       & \textbf{.524} & .520 & \textbf{.524} & .514 & .566 & .564 & .570 & \textbf{.572} & .598 & .604 & .590 & \textbf{.612} \\
\midrule
\multirow{4}{*}{XStoryCloze} & \multirow{4}{*}{es}
  & text      & \textbf{.516} & .508 & .514 & .502 & \textbf{.566} & .552 & .556 & .538 & .566 & .568 & .568 & \textbf{.570} \\
& & romanized & \textbf{.526} & .520 & .520 & .486 & .572 & .576 & .568 & \textbf{.590} & \textbf{.604} & .592 & \textbf{.604} & .592 \\
& & ipa       & \textbf{.528} & .520 & .526 & .520 & .556 & \textbf{.572} & .562 & .566 & .572 & \textbf{.590} & \textbf{.590} & .584 \\
& & text$\rightarrow$rom & .514 & .506 & \textbf{.516} & .492 & \textbf{.550} & .540 & .542 & .542 & .576 & .558 & .566 & \textbf{.580} \\
\midrule
\multirow{4}{*}{XStoryCloze} & \multirow{4}{*}{ru}
  & text      & .494 & \textbf{.498} & .488 & .496 & .556 & \textbf{.560} & \textbf{.560} & .554 & .584 & .580 & .582 & \textbf{.588} \\
& & romanized & \textbf{.534} & .520 & .520 & .494 & .572 & .568 & \textbf{.574} & .562 & .622 & .618 & .622 & \textbf{.624} \\
& & ipa       & .476 & .472 & \textbf{.478} & .460 & \textbf{.518} & .510 & .508 & .514 & .512 & .514 & \textbf{.520} & .516 \\
& & text$\rightarrow$rom & \textbf{.456} & \textbf{.456} & \textbf{.456} & .454 & .458 & .464 & .456 & \textbf{.472} & .448 & .464 & .462 & \textbf{.478} \\
\midrule
\multirow{4}{*}{XStoryCloze} & \multirow{4}{*}{hi}
  & text      & \textbf{.528} & .502 & .504 & .470 & .560 & .570 & .570 & \textbf{.584} & \textbf{.592} & .568 & .568 & .566 \\
& & romanized & .536 & \textbf{.540} & .536 & .526 & .570 & \textbf{.572} & .566 & .558 & \textbf{.590} & .568 & .576 & .580 \\
& & ipa       & \textbf{.506} & .504 & .486 & .476 & .560 & \textbf{.568} & .556 & .564 & .574 & \textbf{.584} & .566 & .576 \\
& & text$\rightarrow$rom & \textbf{.476} & \textbf{.476} & .464 & .462 & .476 & .464 & \textbf{.482} & .456 & .478 & \textbf{.480} & .472 & .470 \\
\midrule
\multirow{4}{*}{XCopa} & \multirow{4}{*}{ta}
  & text      & \textbf{.542} & .540 & .536 & .532 & .566 & .572 & \textbf{.580} & .574 & .586 & .582 & \textbf{.592} & .578 \\
& & romanized & .530 & .524 & \textbf{.534} & .530 & .562 & .562 & .564 & \textbf{.574} & .564 & \textbf{.582} & .580 & .576 \\
& & ipa       & .554 & .556 & .554 & \textbf{.558} & .556 & \textbf{.580} & .576 & .570 & .596 & \textbf{.606} & .580 & .590 \\
& & text$\rightarrow$rom & .564 & \textbf{.570} & \textbf{.570} & .564 & .558 & .554 & .566 & \textbf{.572} & \textbf{.564} & .550 & \textbf{.566} & .550 \\
\bottomrule
\end{tabular}
\caption{Accuracy on XStoryCloze and XCOPA across scales and $k$-shot settings. Bold marks the best score per row within each scale block.}
\label{tab:scale-kshot-results}
\end{table*}

\subsection{Full $k$-shot results}
\label{app:ICL-kshot}

\autoref{tab:scale-kshot-results} reports the same evaluations as the main-text figure across three scales (Small, Medium, Large) and four shot counts ($k{=}0, 1, 2, 4$). Three patterns hold across the full table. First, few-shot demonstrations provide little reliable improvement over zero-shot: within-scale variation across $k$ is small and often non-monotonic, while increasing model scale yields consistent gains for representations trained from scratch in the target surface form. Second, IPA's gap relative to text and romanization persists across $k$ values, indicating that its ICL deficit is a property of the representation rather than an artifact of the zero-shot setting. Third, the text$\rightarrow$rom rows---text-pretrained models evaluated on romanized inputs at inference time---generally underperform the same models evaluated on text, and gain less from added demonstrations or scale than the matched-representation configurations do; an inference-time surface-form mismatch is not rescued by in-context examples.

\subsection{Progression of cross-lingual transfer during pretraining}
\label{app:convergence}

\begin{figure}[t]
  \centering
  \includegraphics[width=\columnwidth]{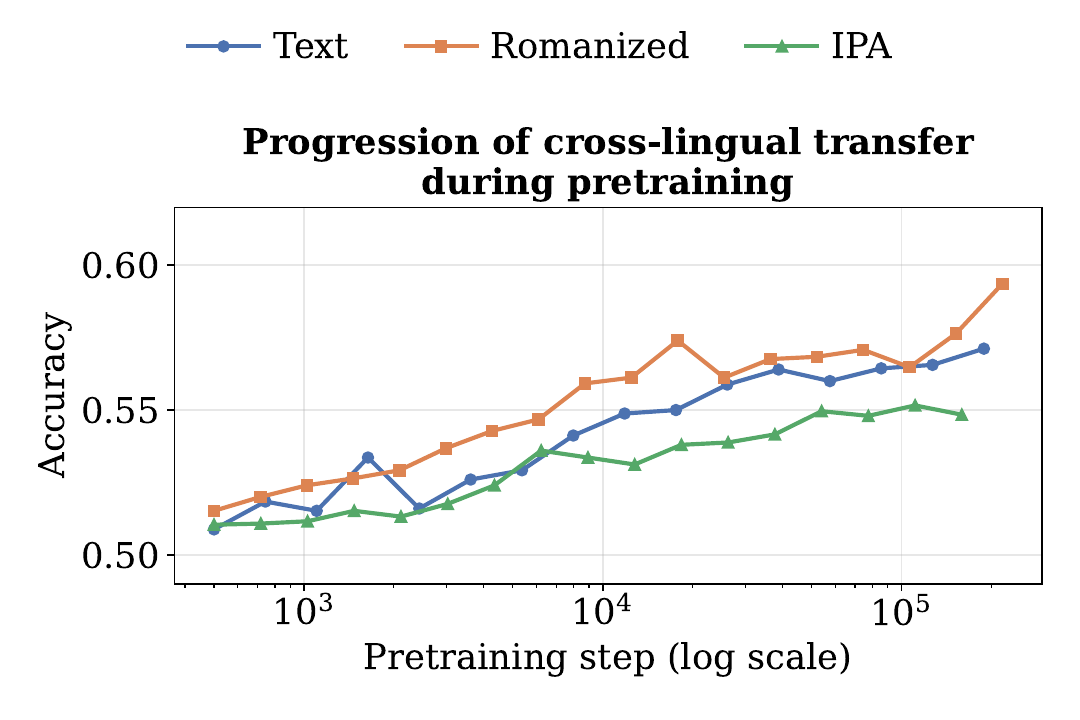}
\caption{Zero-shot cross-lingual transfer accuracy on XStoryCloze and XCOPA at intermediate pretraining checkpoints (medium models, log-scale x-axis).}  \label{fig:convergence}
\end{figure}

To test whether a shared surface form speeds convergence rather than only raising final accuracy, we evaluate cross-lingual transfer at intermediate pretraining checkpoints (\autoref{fig:convergence}). The ordering is stable almost from the outset: romanized $>$ text $>$ IPA at nearly every checkpoint. The gaps do not close as training proceeds. Romanization's advantage is persistent.

\section{Impact of the Tokenizer Integer-Overflow Bug}
\label{app:i64}
All results in the main paper are reported with tokenizers trained using the \texttt{tokenizers} library. The \texttt{tokenizers} library \citep{huggingface_tokenizers} accumulates byte-pair frequencies in a signed 32-bit integer during BPE training, so pairs occurring more than $2^{31}$ times overflow and are dropped or mis-ranked.\footnote{\url{https://github.com/huggingface/tokenizers/issues/2058}} We patched the accumulator to 64-bit, verified byte-identical output below the threshold, and retrained the tokenizers and affected models. Then we re-evaluated on MASSIVE fine-tuning and zero-shot prompting. Re-running the remaining evaluations (XNLI, XL-Sum) was beyond our compute budget.

\begin{table*}
\centering
\scriptsize
\setlength{\tabcolsep}{3pt}
\begin{tabular}{lll ccccccccc}
\toprule
\textbf{Size} & \textbf{Rep.} & \textbf{Tok.}
& \textbf{EN} & \textbf{ES} & \textbf{HI} & \textbf{UR}
& \textbf{RU} & \textbf{PL} & \textbf{TA} & \textbf{ML} & \textbf{Macro} \\
\midrule
\multirow{6}{*}{Small}
& \multirow{2}{*}{Text} & orig
& \textbf{75.56} & \textbf{72.62} & \textbf{71.13} & \textbf{63.32}
& \textbf{73.10} & 71.02 & 65.58 & \textbf{66.16} & \textbf{69.81} \\
& & i64
& 74.38 & 72.41 & 71.04 & 63.17
& 72.36 & \textbf{71.63} & \textbf{66.31} & 65.92 & 69.65 \\
\cmidrule(lr){2-12}
& \multirow{2}{*}{IPA} & orig
& \textbf{76.87} & 74.72 & \textbf{74.08} & 71.63
& 71.75 & \textbf{72.68} & 70.17 & \textbf{69.21} & 72.64 \\
& & i64
& 76.69 & \textbf{75.38} & 73.27 & \textbf{72.51}
& \textbf{71.84} & 72.42 & \textbf{70.91} & 68.57 & \textbf{72.70} \\
\cmidrule(lr){2-12}
& \multirow{2}{*}{Romanized} & orig
& 79.90 & \textbf{77.80} & 75.70 & \textbf{73.70}
& 77.80 & \textbf{77.40} & 69.80 & 71.90 & 75.50 \\
& & i64
& \textbf{80.73} & 77.62 & \textbf{75.91} & 73.48
& \textbf{78.46} & 76.35 & \textbf{69.92} & \textbf{72.71} & \textbf{75.65} \\
\midrule
\multirow{6}{*}{Medium}
& \multirow{2}{*}{Text} & orig
& 81.98 & \textbf{80.31} & 80.60 & 75.31
& \textbf{80.95} & 78.49 & 77.64 & 80.05 & 79.42 \\
& & i64
& \textbf{82.33} & 79.14 & \textbf{80.71} & \textbf{75.38}
& 80.21 & \textbf{79.26} & \textbf{78.17} & \textbf{80.84} & \textbf{79.51} \\
\cmidrule(lr){2-12}
& \multirow{2}{*}{IPA} & orig
& 82.77 & \textbf{79.19} & 80.97 & \textbf{78.68}
& 80.10 & \textbf{80.65} & 76.57 & 79.15 & 79.76 \\
& & i64
& \textbf{83.61} & 79.04 & \textbf{81.02} & 77.91
& \textbf{81.03} & 79.84 & \textbf{77.26} & \textbf{79.68} & \textbf{79.92} \\
\cmidrule(lr){2-12}
& \multirow{2}{*}{Romanized} & orig
& 83.04 & 81.40 & 80.06 & \textbf{77.36}
& \textbf{82.54} & 81.45 & 76.84 & \textbf{78.21} & 80.11 \\
& & i64
& \textbf{83.17} & \textbf{81.56} & \textbf{80.14} & 76.47
& 82.38 & \textbf{82.13} & \textbf{77.51} & 77.83 & \textbf{80.15} \\
\midrule
\multirow{6}{*}{Large}
& \multirow{2}{*}{Text} & orig
& 83.99 & \textbf{82.21} & 84.67 & 79.89
& 83.59 & \textbf{82.95} & 82.28 & 83.96 & 82.94 \\
& & i64
& \textbf{85.06} & 81.48 & \textbf{85.22} & \textbf{80.94}
& \textbf{84.17} & 82.26 & \textbf{83.61} & \textbf{84.44} & \textbf{83.40} \\
\cmidrule(lr){2-12}
& \multirow{2}{*}{IPA} & orig
& \textbf{86.75} & \textbf{83.96} & \textbf{85.81} & 82.48
& 84.80 & \textbf{84.57} & 82.55 & \textbf{84.67} & \textbf{84.45} \\
& & i64
& 86.21 & 83.76 & 85.31 & \textbf{83.15}
& 84.80 & 84.06 & 82.55 & 84.40 & 84.28 \\
\cmidrule(lr){2-12}
& \multirow{2}{*}{Romanized} & orig
& 86.42 & \textbf{85.88} & \textbf{86.28} & \textbf{82.72}
& \textbf{86.28} & \textbf{85.27} & 82.28 & 84.30 & \textbf{84.93} \\
& & i64
& \textbf{86.45} & 84.77 & 85.51 & 82.35
& 85.78 & 84.87 & \textbf{82.41} & \textbf{85.24} & 84.67 \\
\bottomrule
\end{tabular}
\caption{Fine-tuning F1 (\%) on MASSIVE with the original (\emph{orig}) and i64-fixed (\emph{i64}) tokenizers. Bold marks the better setting per language and scale; \emph{Macro} averages the eight languages.}
\label{tab:seen-finetuning-massive-orig-vs-i64}
\end{table*}

\begin{table}[!t]
\centering
\scriptsize
\setlength{\tabcolsep}{2pt}
\begin{tabular}{lll cccccc}
\toprule
& & & \multicolumn{2}{c}{\textbf{Small}} & \multicolumn{2}{c}{\textbf{Medium}} & \multicolumn{2}{c}{\textbf{Large}} \\
\cmidrule(lr){4-5} \cmidrule(lr){6-7} \cmidrule(lr){8-9}
\textbf{Task} & \textbf{Lang} & \textbf{Repr.} & orig & i64 & orig & i64 & orig & i64 \\
\midrule
\multirow{3}{*}{XSC} & \multirow{3}{*}{en}
  & text & \textbf{.532} & .528 & \textbf{.594} & .588 & \textbf{.600} & .584 \\
& & rom. & \textbf{.546} & .542 & .614 & .614 & \textbf{.640} & .634 \\
& & ipa & .524 & \textbf{.528} & \textbf{.566} & .562 & \textbf{.598} & .588 \\
\midrule
\multirow{3}{*}{XSC} & \multirow{3}{*}{es}
  & text & \textbf{.516} & .508 & .566 & \textbf{.572} & \textbf{.566} & .560 \\
& & rom. & \textbf{.526} & .522 & \textbf{.572} & .568 & \textbf{.604} & .600 \\
& & ipa & \textbf{.528} & .518 & \textbf{.556} & .546 & .572 & .572 \\
\midrule
\multirow{3}{*}{XSC} & \multirow{3}{*}{ru}
  & text & .494 & \textbf{.506} & .556 & \textbf{.562} & .584 & \textbf{.586} \\
& & rom. & .534 & \textbf{.536} & .572 & \textbf{.578} & .622 & .622 \\
& & ipa & \textbf{.476} & .470 & \textbf{.518} & .516 & .512 & \textbf{.522} \\
\midrule
\multirow{3}{*}{XSC} & \multirow{3}{*}{hi}
  & text & \textbf{.528} & .526 & .560 & \textbf{.570} & \textbf{.592} & .584 \\
& & rom. & .536 & \textbf{.542} & \textbf{.570} & .562 & \textbf{.590} & .586 \\
& & ipa & .506 & \textbf{.516} & \textbf{.560} & .554 & .574 & \textbf{.582} \\
\midrule
\multirow{3}{*}{XCOPA} & \multirow{3}{*}{ta}
  & text & \textbf{.542} & .530 & \textbf{.566} & .558 & \textbf{.586} & .582 \\
& & rom. & .530 & \textbf{.552} & .562 & .562 & .564 & .564 \\
& & ipa & .554 & \textbf{.564} & .556 & \textbf{.564} & \textbf{.596} & .592 \\
\midrule
\multirow{3}{*}{\textbf{Avg}} & \multirow{3}{*}{--}
  & text & \textbf{.522} & .520 & .568 & \textbf{.570} & \textbf{.586} & .579 \\
& & rom. & .534 & \textbf{.539} & \textbf{.578} & .577 & \textbf{.604} & .601 \\
& & ipa & .518 & \textbf{.519} & \textbf{.551} & .548 & .570 & \textbf{.571} \\
\bottomrule
\end{tabular}
\caption{Zero-shot accuracy on XStoryCloze (XSC) and XCOPA (ta) with the original (\emph{orig}) and i64-fixed (\emph{i64}) tokenizers. Bold marks the better setting per scale; \emph{Avg} averages the five task--language cells.}
\label{tab:zeroshot-orig-vs-i64}
\end{table}

Tables~\ref{tab:seen-finetuning-massive-orig-vs-i64}
and~\ref{tab:zeroshot-orig-vs-i64} compare the two settings. MASSIVE macro-F1 changes by at most $0.46$ points and zero-shot average accuracy by at most $0.007$. In both cases, the changes are significantly smaller than the gaps between the representations. Every pairwise ordering of the three representations is preserved at every scale. 
\begin{table}[!t]
\centering
\small
\setlength{\tabcolsep}{5pt}
\begin{tabular}{lcccccc}
\toprule
& \multicolumn{2}{c}{\textbf{Text}} & \multicolumn{2}{c}{\textbf{IPA}} & \multicolumn{2}{c}{\textbf{Romanized}} \\
\cmidrule(lr){2-3} \cmidrule(lr){4-5} \cmidrule(lr){6-7}
\textbf{Pair} & orig & i64 & orig & i64 & orig & i64 \\
\midrule
Tam--Mal & 0.004 & 0.004 & 0.127 & 0.125 & 0.195 & 0.191 \\
Rus--Pol & 0.066 & 0.067 & 0.104 & 0.103 & 0.200 & 0.199 \\
Eng--Spa & 0.149 & 0.149 & 0.059 & 0.064 & 0.154 & 0.153 \\
Hin--Urd & 0.007 & 0.007 & 0.243 & 0.231 & 0.066 & 0.065 \\
\bottomrule
\end{tabular}
\caption{Frequency-weighted subword overlap within each language pair under the original (\emph{orig}) and i64-fixed (\emph{i64}) tokenizers.}
\label{tab:i64-jaccard}
\end{table}

Cross-lingual transfer is stable as well. Frequency-weighted subword overlap within each language pair, computed as described in \autoref{app:overlap-details}, changes at most $0.012$ (\autoref{tab:i64-jaccard}).

Therefore, our conclusions are unchanged under either setting.

\section{AI Disclosure}
We used generative AI tools for limited writing and coding assistance, including grammar, clarity, organization suggestions, and debugging support. All research ideas, experimental design decisions, code, results, analysis, and final text were reviewed and verified by the authors.

\end{document}